\documentclass[letterpaper]{article} 
\usepackage{style}  
\usepackage{times}  
\usepackage{helvet}  
\usepackage{courier}  
\usepackage[hyphens]{url}  
\usepackage{graphicx} 
\usepackage{natbib}  
\usepackage{caption} 
\usepackage{algorithm}
\usepackage{algorithmic}

\usepackage{newfloat}
\usepackage{listings}
\DeclareCaptionStyle{ruled}{labelfont=normalfont,labelsep=colon,strut=off} 
\floatstyle{ruled}
\newfloat{listing}{tb}{lst}{}
\floatname{listing}{Listing}
\title{Text2CAD: Text to 3D CAD Generation via Technical Drawings}
\author{Mohsen Yavartanoo$^{1}$ \qquad Sangmin Hong$^{2}$ \qquad Reyhaneh Neshatavar$^{1}$ \qquad Kyoung Mu Lee$^{1,2}$ \\$^{1}$Dept. of ECE \& ASRI, $^{2}$IPAI, Seoul National University, Seoul, Korea\\
{\tt\small \{myavartanoo,mchiash2,reyhanehneshat,kyoungmu\}@snu.ac.kr}}

\usepackage{subfig}
\usepackage{amsmath}
\usepackage{amssymb}
\usepackage{booktabs}
\usepackage{pifont}
\newcommand{\cmark}{\ding{51}}%
\newcommand{\xmark}{\ding{55}}%

\usepackage{bibentry}

\begin{document}
\def\eg{\emph{e.g.}}
\def\ie{\emph{i.e.}}
\def\etal{\emph{et al.}}

\maketitle

\begin{abstract}
The generation of industrial Computer-Aided Design~(CAD) models from user requests and specifications is crucial to enhancing efficiency in modern manufacturing. 
Traditional methods of CAD generation rely heavily on manual inputs and struggle with complex or non-standard designs, making them less suited for dynamic industrial needs. 
To overcome these challenges, we introduce Text2CAD, a novel framework that employs stable diffusion models tailored to automate the generation process and efficiently bridge the gap between user specifications in text and functional CAD models. 
This approach directly translates the user's textural descriptions into detailed isometric images, which are then precisely converted into orthographic views, \eg, top, front, and side, providing sufficient information to reconstruct 3D CAD models.
This process not only streamlines the creation of CAD models from textual descriptions but also ensures that the resulting models uphold physical and dimensional consistency essential for practical engineering applications. 
Our experimental results show that Text2CAD effectively generates technical drawings that are accurately translated into high-quality 3D CAD models, showing substantial potential to revolutionize CAD automation in response to user demands. 
\end{abstract}

\section{Introduction}
\label{sec:Text2CAD_intro}
%

Industrial Computer-Aided Design~(CAD) models are essential tools in modern manufacturing. 
They serve as detailed blueprints for a wide range of products, from simple tools to complex machinery. 
CAD software enables engineers and designers to create precise geometric representations of products, facilitating visualization, simulation, and manufacturing. 
Automating the generation of these models from conceptual descriptions not only enhances productivity and reduces time-to-market but also fosters innovation.
This aligns with broader industry trends toward digital transformation and smart manufacturing.

Despite progress, automating the generation of CAD from user requests poses challenges~\cite{kasik2005ten}. 
Traditional methods, often manual and time-consuming, struggle with complex designs and fail to handle the nuances required by modern specifications.

Recent advances in diffusion models have shown the potential to generate detailed images from textual prompts~\cite{zhao2023unleashing, Textdiffuser, Dreambooth, Attend-and-excite, zhang2023adding}. 
However, these models typically do not grasp three-dimensional constraints, leading to outputs that, while visually impressive, fall short in practical engineering applications. 
CAD models, on the other hand, can be effectively represented through technical drawings, which are 2D projections of the 3D model to orthographic views~\cite{governi20133d}. 
Although diffusion models excel at generating these 2D images, they often do not maintain the necessary physical and dimensional consistency required for these drawings to be directly usable in manufacturing. 
This inconsistency is a significant limitation, as each drawing must accurately reflect the physical properties of the CAD model across multiple orthographic views to be useful in practical applications.

\begin{figure}[!t]
\centering
\subfloat{
    \includegraphics[width=\linewidth]{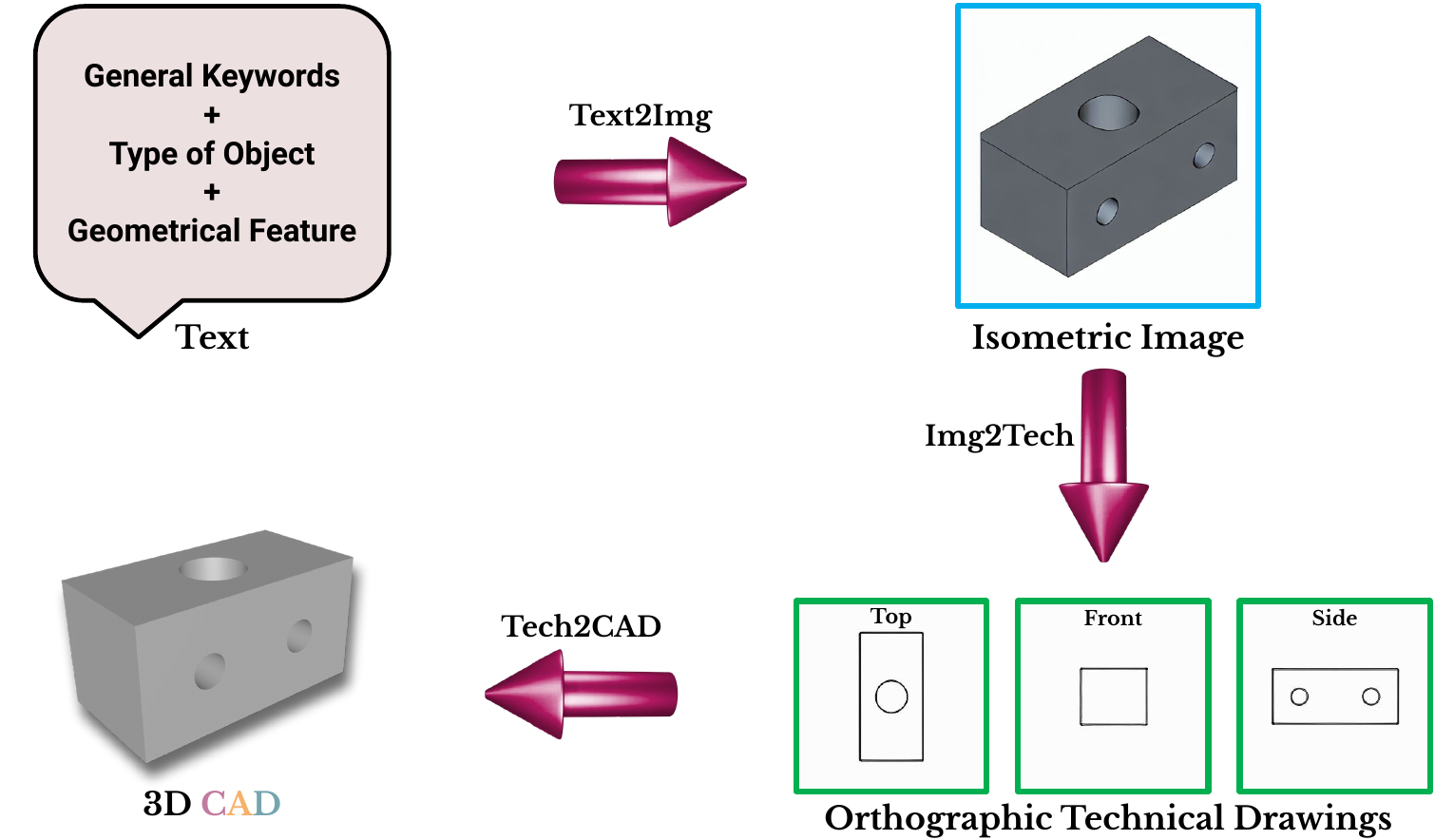}}
    \caption{\textbf{Overview of our method.}  Our method converts textual descriptions into a 3D CAD model through a multi-step process. First, the text is transformed into an isometric image representing the described features. This image is then mapped into orthographic technical drawings, which serve as the foundation for generating the 3D CAD model.}
    \label{fig:Text2CAD_teaser}
\end{figure}

In response to these automation challenges, we develop Text2CAD, a framework that leverages isometric drawings and stable diffusion models to bridge the gap between textual descriptions and precise CAD models, as shown in Figure~\ref{fig:Text2CAD_teaser}. 
Given a user's textural description of a 3D object, our method starts by generating an isometric image that effectively captures the geometric features of the 3D object across various perspectives. 
Accordingly, a novel view generation difMohsefusion model takes the isometric image as input and transforms it into detailed orthographic views. 
This approach not only streamlines the CAD generation process, but also ensures the accuracy and consistency needed for practical manufacturing applications. 
By enabling the direct conversion of text descriptions into comprehensive technical drawings, Text2CAD significantly enhances the efficiency and accessibility of CAD model creation, aligning with the demands of modern industry.

To facilitate robust training and evaluation of our model,
we introduce a new dataset comprising detailed descriptions and corresponding technical drawings of CAD models. This dataset serves as a foundation for training and benchmarking the Text2CAD framework. 
%
Our experimental results confirm that the Text2CAD framework reliably produces technical drawings that are accurately translated into practical 3D CAD models. 
This reveals the capability of our framework to bridge the gap between textual descriptions and functional engineering output, effectively streamlining the CAD model generation.
%
Our main contributions are threefold:
\begin{itemize}
    \item We introduce Text2CAD, a novel framework that uses stable diffusion models to automate the creation of CAD models from textual descriptions.

    \item Our method streamlines the process by generating a detailed isometric drawing and transforming it into consistent orthographic views, \eg, top, front, and side.

    \item Experimental results demonstrate that Text2CAD reliably produces technical drawings that translate into high-quality 3D CAD models, effectively bridging the gap between textual prompts and practical engineering outputs.
\end{itemize}

\section{Related work}
\label{sec:related_selfwarp}
In this section, we review prior studies pertinent to our method, divided into CAD generation and diffusion models. Each category is critical to understanding the advancements in our field and the niche that our research aims to fill.

\subsection{CAD generation}
Traditionally, CAD systems largely relied on manual drawing techniques, which required extensive user input and expertise~\cite{tovey1989drawing, mclaren2008exploring, ibrahim2010comparison, amadori2012flexible}. 
These early systems were primarily vector-based drawing tools that facilitated the design of 2D and basic 3D models. 
As technology advanced, CAD software saw significant enhancements, evolving into more sophisticated platforms capable of detailed 3D modeling and simulation~\cite{ibrahim2010comparison}. 
These advancements transformed how professionals in engineering, architecture, and manufacturing approached the design process. 
With the advent of deep learning, the capability of CAD systems expanded further, leading to the exploration of automated algorithms for generating complex 3D models~\cite{lee2022dataset}. 
Deep learning-based methods in CAD generation have effectively utilized diverse observations such as point clouds, voxels, and images to create corresponding 3D CAD models. 
These techniques have revolutionized the CAD design process, enabling automated and precise model generation from straightforward observations of objects, whether through images or other 3D representations~\cite{Capri-net, Extrudenet, Csg-stump}.
However, despite these advancements, the challenge of simplifying the manufacturing process remains. 
To address this, our research aims to innovate further by enabling the generation of CAD models directly from text prompts provided by users. 
This would streamline the design-to-production pipeline and make CAD generation more accessible to non-experts.

\subsection{Diffusion models}
The concept of a diffusion process in data science is derived from the statistical mechanics principle~\cite{schmittmann1998driven}, in which particles move from areas of higher concentration to lower concentration until equilibrium is reached. 
In machine learning, diffusion models~\cite{cao2024survey} have been developed as a class of generative models that learn to reverse this process. 
These models start with a random noise distribution and gradually learn to subtract this noise to recreate data samples from the learned distribution. 
One of the groundbreaking advancements in this area is the development of Stable Diffusion~\cite{stablediffusion}, a model that has made it feasible to generate detailed images from textual prompts. 
This capability stems from the model's deep understanding of the data's latent space, allowing it to generate highly detailed and specific images based on simple text descriptions.
Applications of diffusion models have been particularly impressive in areas requiring high-fidelity visual outputs. 
In novel view generation, these models provide new perspectives of objects or scenes from minimal initial viewpoints~\cite{Zero-1-to-3}, significantly enhancing virtual reality and 3D animation processes. 
Another vital application is 3D shape reconstruction, where diffusion models interpret various data inputs to recreate complex three-dimensional shapes, proving essential in automated design and manufacturing processes.
Leveraging the generative capabilities of diffusion models, our research utilizes these models to create precise technical drawings from textual descriptions. 
By harnessing this technology, we aim to bridge the gap between text-based inputs and detailed 3D CAD models, making the process of generating CAD designs both more accessible and efficient.

\section{Method}
\label{sec:method_Text2CAD}
%
\begin{figure*}[!h]
\centering
\subfloat{
    \includegraphics[width=\linewidth]{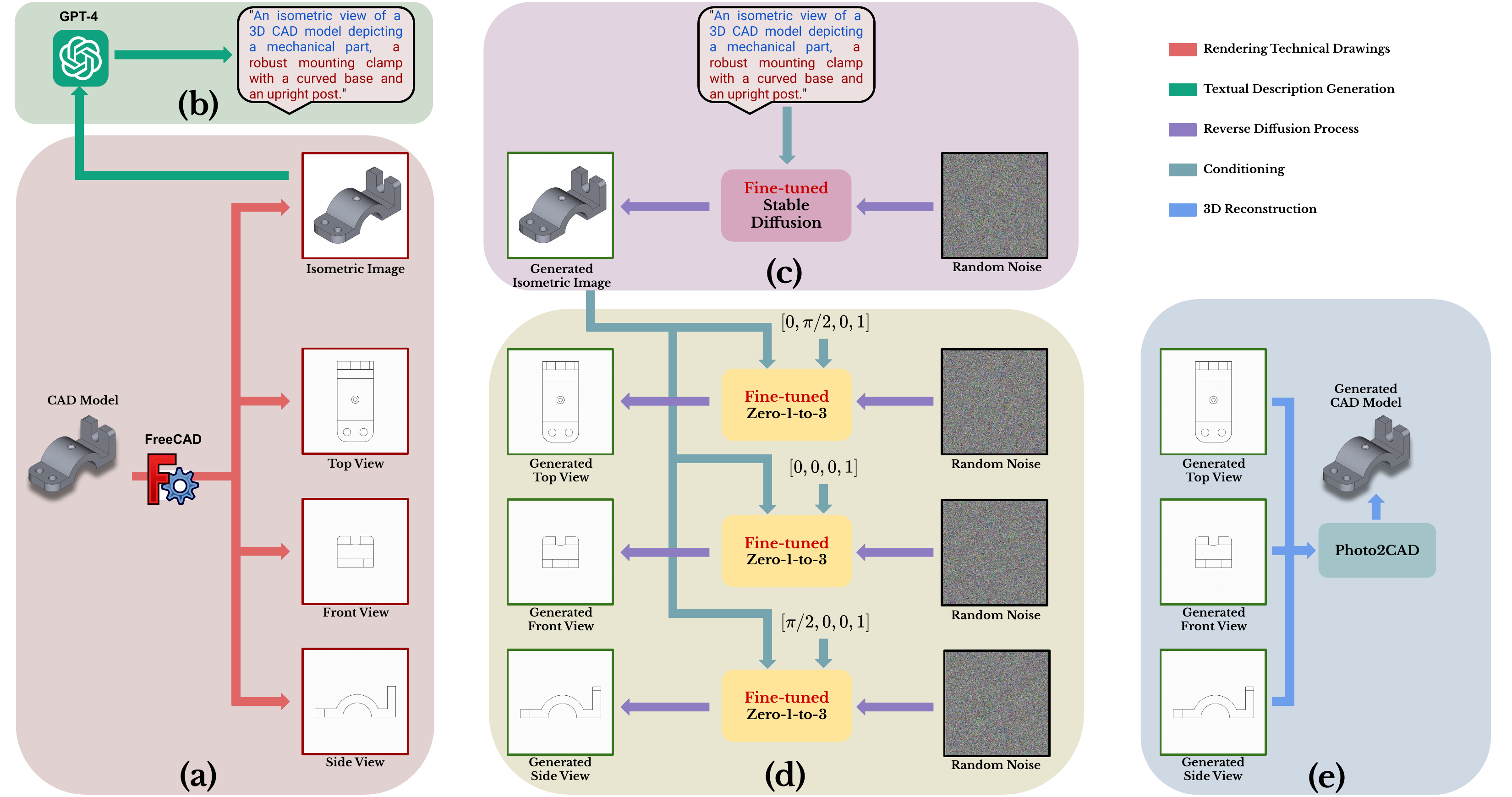}}
    \caption{\textbf{Overall view of our Text2CAD framework.} The dataset creation process involves (a) generating technical drawings from a CAD model and (b) producing textual descriptions of the CAD model using GPT-4. The CAD generation process includes (c) generating isometric images based on the textual descriptions, (d) deriving orthographic technical drawings from the generated isometric images, and finally, (e) reconstructing the CAD model from the generated orthographic technical drawings.}
    \label{fig:Text2CAD_framework}
\end{figure*}

In this section, we present our method for generating CAD models from textual descriptions, as shown in Figure~\ref{fig:Text2CAD_framework}. 
We begin with an overview of CAD models and technical drawings and their role in engineering and manufacturing. 
Next, we describe the process of creating our dataset as a key step in generating 3D CAD models. 
We then discuss our model training approaches to improve drawing accuracy and consistency, followed by an explanation of the tools and techniques used to transform 2D drawings into 3D CAD models.

\subsection{Technical drawings}\label{sec:techdraw}
A Computer-Aided Design~(CAD) model is a digital representation used in engineering, architecture, and manufacturing for precise visualization and analysis. 
Created with CAD software, these models improve design accuracy and production efficiency by detailing an object's geometry and features in 3D formats or 2D technical drawings.

Technical drawings are crucial in engineering, as they specify component designs, dimensions, and manufacturing requirements. 
They follow standards like ANSI Y14.5 and ISO 8015 for precision, using orthogonal projections~(ISO 128, ASME Y14.3) and universal symbols~(ISO 1302) to ensure clarity and accuracy. 
CAD software such as AutoCAD, SolidWorks, CATIA, and FreeCAD integrates these standards to maintain consistency and facilitate effective communication in manufacturing.

\subsection{Dataset}
We detail our creation of a unique dataset consisting of technical drawings and corresponding textual descriptions for 3D CAD models.
This comprehensive dataset is designed to support the development and testing of machine learning models, particularly in automating the generation and interpretation of CAD designs.

\subsubsection{Rendering technical drawing}
\begin{figure}[!h]
\centering
\subfloat{       
\includegraphics[width=0.42\linewidth, page=2]{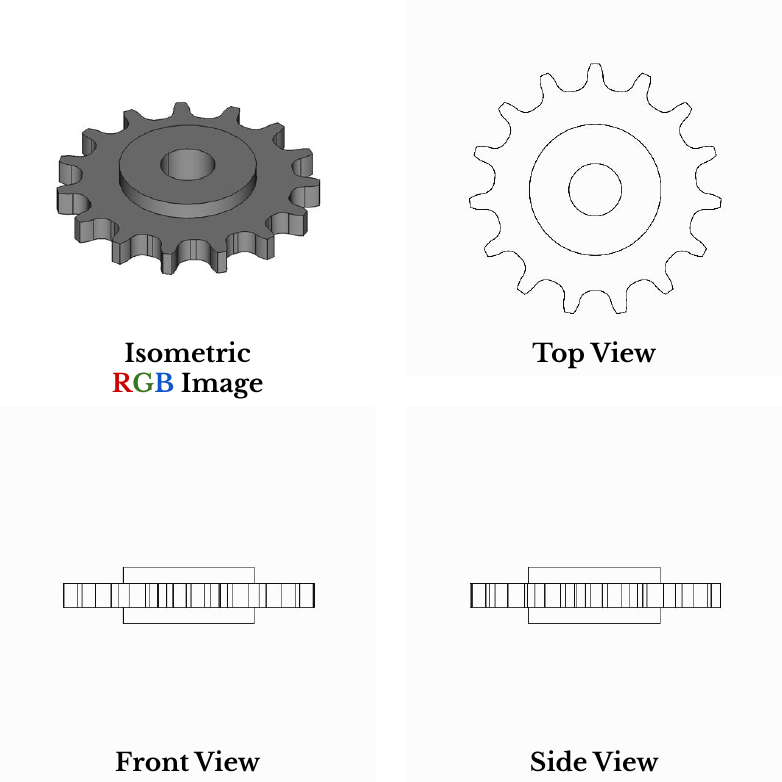}}
\hfill
\subfloat{
\includegraphics[width=0.42\linewidth, page=4]{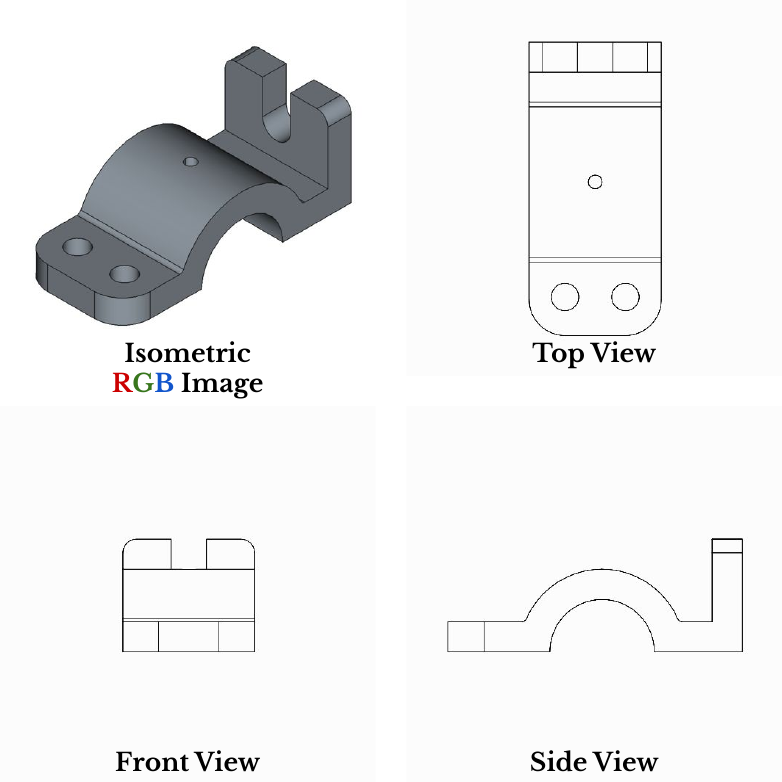}}
    \caption{\textbf{Rendered technical drawings.} Isometric images and the orthographic technical drawings of the CAD models.}
    \label{fig:Text2CAD_techdraw}
\end{figure}

We utilize FreeCAD to automate the rendering of both isometric images and orthographic technical drawings \eg, front, top, and side, that are necessary for comprehensive technical documentation. 
The process begins with importing STEP files, which are standardized 3D model formats widely used in the industry. 
These models are scaled and manipulated to align with the standard drawing templates provided by the TechDraw workbench in FreeCAD. 
Figure~\ref{fig:Text2CAD_techdraw} shows the rendered isometric image and orthographic technical drawings samples.

\begin{figure}[!h]
\centering
\subfloat{
    \includegraphics[width=\linewidth]{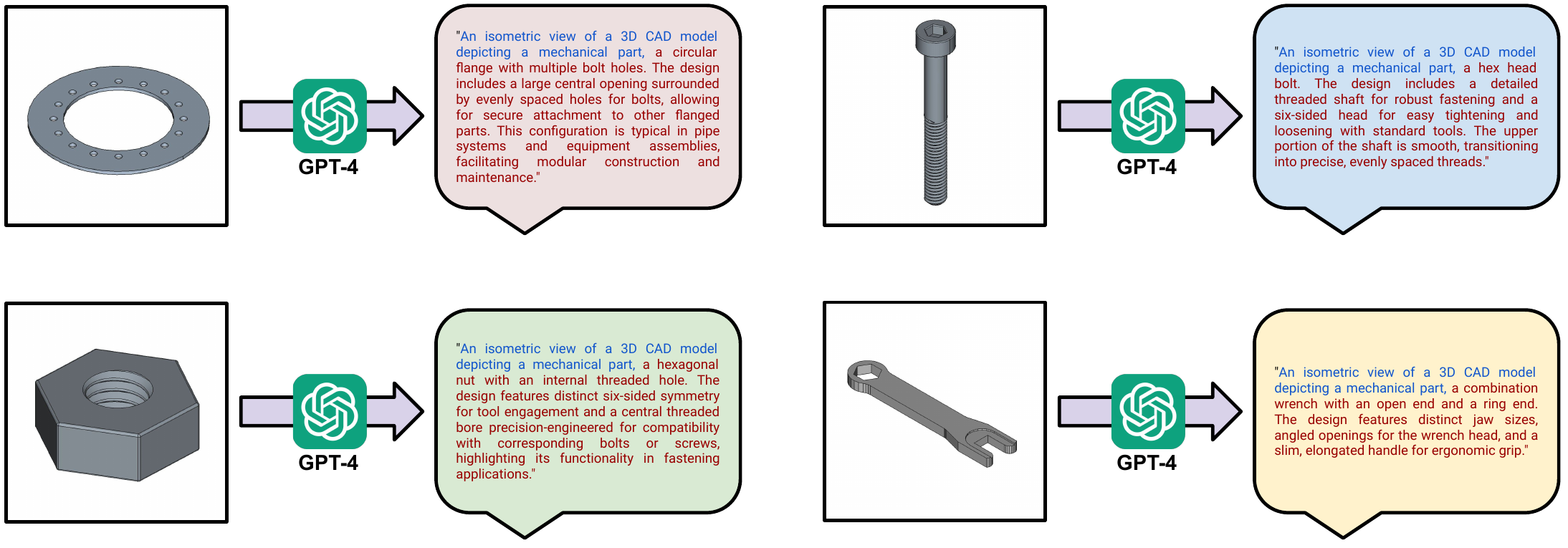}}
    \caption{\textbf{Generated descriptions.} GPT-4 receives isometric images and provides descriptions within a template. }
    \label{fig:Text2CAD_gpt4}
\end{figure}

\subsubsection{Textual description generation}
Accurate textual descriptions are crucial for effective text-to-CAD generation, yet manually labeling CAD models is difficult due to their complexity and specialized terminology. 
Each CAD model requires precise descriptions that capture its intricate details and functionality. 
To address this, we use GPT-4, an advanced language model adept at interpreting complex visuals and generating contextually relevant text, making it ideal for describing CAD models as shown in Figure~\ref{fig:Text2CAD_gpt4}.
We develop a template to guide GPT-4 in generating descriptions starting with "{\it An isometric view of a 3D CAD model depicting a mechanical part,}" and including details on the object’s type and key features.
By inputting the isometric image into GPT-4, we ensure the descriptions align closely with the visual data.
Using GPT-4, we efficiently produce detailed and accurate descriptions essential for our text-to-CAD applications, enhancing our dataset and improving the scalability and effectiveness of our CAD generation pipeline.

\subsection{Text to isometric image}\label{sec:Text2CAD_fine} 
\begin{figure}[!h]
\centering
\subfloat{
    \includegraphics[width=0.23\linewidth]{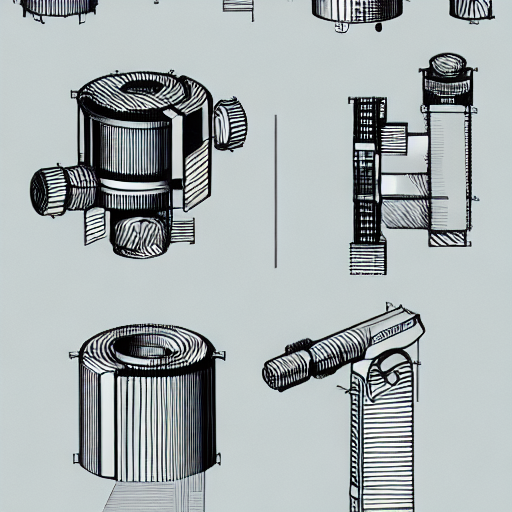}}
\hfill
\subfloat{       
\includegraphics[width=0.23\linewidth]{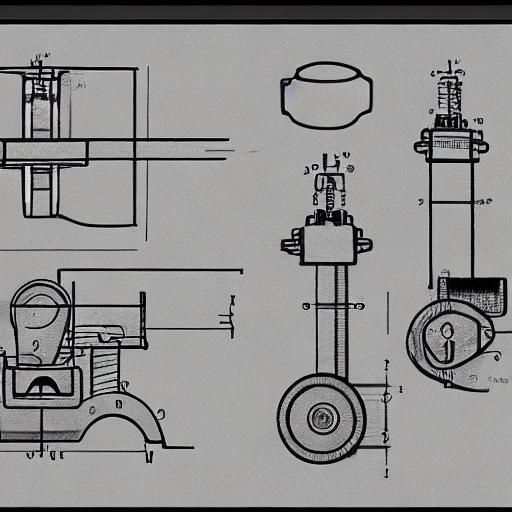}}
\hfill
\subfloat{        
\includegraphics[width=0.23\linewidth]{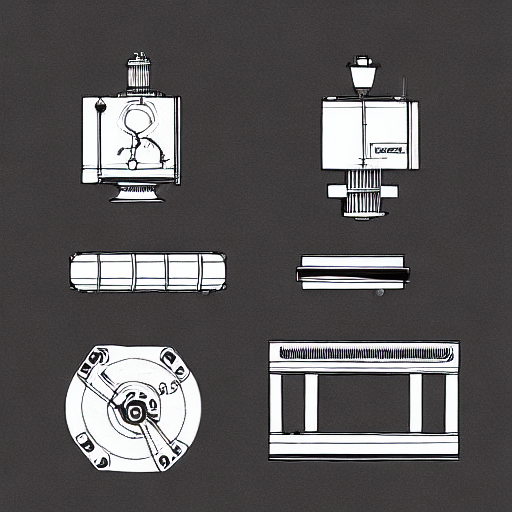}}
\hfill
\subfloat{
\includegraphics[width=0.23\linewidth]{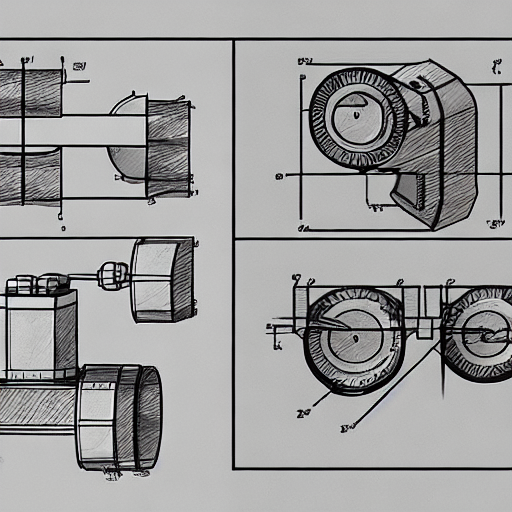}}
    \caption[Technical drawings by a stable diffusion model]{\textbf{Technical drawings by a stable diffusion model.} The images are generated by Stable Diffusion v1-5.}
    \label{fig:Text2CAD_techdiff}
\end{figure}

Stable diffusion models have shown notable proficiency in generating images from textual prompts, effectively producing technical drawings of industrial components as demonstrated in Figure~\ref{fig:Text2CAD_techdiff}. 
However, these models often struggle with maintaining physical consistency across multiple perspectives of the same object, leading to discrepancies and misalignments between different views \eg, top, front, and side, which are critical for a holistic understanding and manufacturing of industrial components.
%
%

\begin{figure}[!h]
\centering
\subfloat{
    \includegraphics[width=\linewidth]{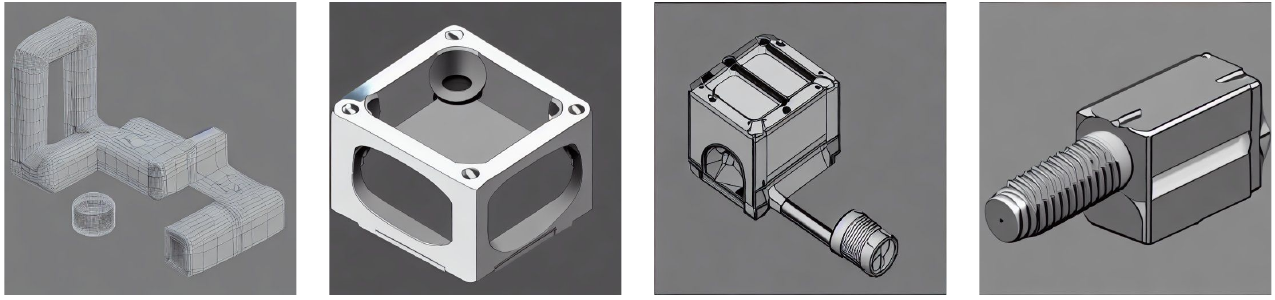}}
    \caption{\textbf{Isometric image by a stable diffusion.} The images are generated by Stable Diffusion v1-5 given text prompt \textit{"An isometric view of a 3D CAD model depicting a mechanical part"}.}
    \label{fig:Text2CAD_difffails}
\end{figure}

To address this limitation, we adopt an alternative approach by initially generating an isometric view from text prompts using diffusion models. 
This isometric view combines all three critical perspectives \eg, top, front, and side, providing a comprehensive representation of the object.
Such a view is highly useful in CAD modeling, effectively conveying intricate details of the object. 
This isometric image then serves as a precursor to generate orthographic images, laying a versatile and informative foundation for further technical illustration. 
Despite their effectiveness in generating contextually relevant images, as depicted in Figure~\ref{fig:Text2CAD_difffails}, pre-trained diffusion models often lack the precision necessary to depict intricate details, precise edges, and surfaces needed for accurate orthographic drawings.
%
%
Accordingly, we fine-tune a stable diffusion~\cite{stablediffusion} model to learn the process of generating rendered images $I$ from an isometric view, given a generated text prompt $P$ describing the target object by minimizing the following objective:
\begin{equation}
    \mathcal{L}_{iso} = \min_{\theta_1} \mathbb{E}_{\substack{z \sim \mathcal{E}(I), t, \\ \epsilon \sim \mathcal{N}(0,1)}} \left\| \epsilon - \epsilon_{\theta_1}(z_t, t, c(P)) \right\|^2_2,
\end{equation}
where the parameters of the U-Net module $\epsilon_{\theta_1}$ are trainable, while the text encoder $c$ is considered with frozen parameters. 
After the model is fine-tuned, during the inference, we can generate images from the isometric view of CAD models by performing iterative denoising from a Gaussian noise image conditioned on text prompt $P$.

\subsection{Isometric to orthographic technical drawings} 
After generating an isometric drawing using the fine-tuned stable diffusion model conditioned on a specified prompt, the subsequent step involves producing orthographic drawings essential for reconstructing a 3D CAD model. 
Utilizing the isometric drawing as a foundational representation, we facilitate the generation of technical drawings from various perspectives such as top, front, and side views. 
This method ensures geometric and physical consistency across views, providing accurate modeling and analysis.

Humans can intuitively convert an isometric image into corresponding orthographic technical drawings using their understanding of perspective and geometric principles. 
However, automating this process requires embedding a comparable level of understanding into a model. 
We achieve this by training the model on paired data consisting of isometric images and their corresponding orthographic technical drawings. 
Through this training, the model discerns geometric relationships between different views and projects the isometric representation onto the orthographic plane efficiently. 
Leveraging advanced machine learning techniques enables us to bridge the gap between human intuition and automated processing, enhancing the generation of accurate orthographic technical drawings from isometric images.

Diffusion models are particularly adept at tasks involving novel view generation and are capable of generating realistic and diverse images from specified input conditions. 
Our task centers around novel view generation conditioned by an isometric image, making diffusion models ideal for our purposes. 
We utilize zero-1-to-3~\cite{Zero-1-to-3}, a diffusion-based method tailored for novel view generation tasks. 
This method excels at generating novel view images given a single image as the condition, along with the relative camera pose, making it well-suited for our task of producing orthographic technical drawings from isometric views.

However, directly applying this method without fine-tuning might not fully meet our requirements, as it typically produces colored rendered images rather than the technical drawings needed for CAD. 
Therefore, we fine-tune this model on our generated dataset to adapt it for generating technical drawings, which have distinct characteristics from rendered images. 
During the fine-tuning process, we maintain the isometric images as the condition of the diffusion model, with the target being the orthographic technical drawings.
The fine-tuning objective is defined as follows:
\begin{equation}
    \mathcal{L}_{tech} = \min_{\theta_2} \mathbb{E}_{\substack{z \sim \mathcal{E}(I), t, \\\epsilon \sim \mathcal{N}(0,1)}} \left\| \epsilon - \epsilon_{\theta_2}(z_t, t, c(I,R,T))) \right\|^2_2,
\end{equation}
where $c$  is CLIP~\cite{clip}, encoding the input isometric image $I$ concatenated with the camera pose parameters $(R,T)$. 
Once the model $\epsilon_{\theta_2}$ is fine-tuned, we can generate the desired orthographic technical drawings given a single isometric image and the corresponding camera pose parameters $(R,T)$, ensuring that all generated views are consistent and accurately represent the same CAD object.

\subsection{Orthographic technical drawings to 3D CAD}
Orthographic technical drawings contain the essential information for reconstructing 3D CAD models, but extracting accurate paths from these images is challenging without an SVG representation. 
Accordingly, we use Photo2CAD~\cite{Photo2CAD}, a straightforward, non-learnable tool built on OpenCV~\cite{opencv}. 
Despite its simplicity, it effectively extracts paths from orthographic images, enabling accurate CAD model reconstruction while preserving the integrity of the original designs.

\section{Experiments}
\label{sec:experiments_Text2CAD}
In this section, we outline our experimental setup, detailing the procedures for generating CAD models from textual prompts. 
We start by describing the dataset preparation for training our deep learning models. 
Then, we cover the technical aspects of our implementation, including the fine-tuned diffusion models and CAD reconstruction. 
We present experiments to test the effectiveness of our approach in producing accurate technical drawings and CAD models from text, followed by ablation studies to assess the robustness and performance of our method.

\subsection{Implementation details}
We employ the ABC dataset~\cite{Abc}, a large-scale collection of one million CAD models including an extensive range of mechanical and industrial shapes, for our experiments. 
Focusing on single-component objects, we select a subset from the last chunk of the dataset, comprising $100,000$ samples in "{\it .step}" format.

To render isometric images, the viewpoint is set at a 45-degree angle above the horizontal plane combined with a 45-degree rotation around the vertical axis. 
This perspective provides a clear and comprehensive view of the three-dimensional structure of the object without distortion.

In creating these technical drawings, a uniform scaling approach is used to ensure consistency across all objects. 
Each object is scaled so that its longest edge measures precisely $2$ units. 
This standardization facilitates easier comparison and comprehension, enhancing the understanding of the depicted elements.
The technical drawings are initially formatted as "{\it .svg}" files, which represent vector paths and their attributes. 
However, directly using these files presents challenges, so they are rendered into images. 
All images are then cropped to a uniform size of $512\times 512$ pixels.
To generate textual descriptions, $1000$ random samples were input into GPT-4, and their corresponding descriptions were collected.

\begin{figure}
\centering
\subfloat{
    \includegraphics[width=\linewidth]{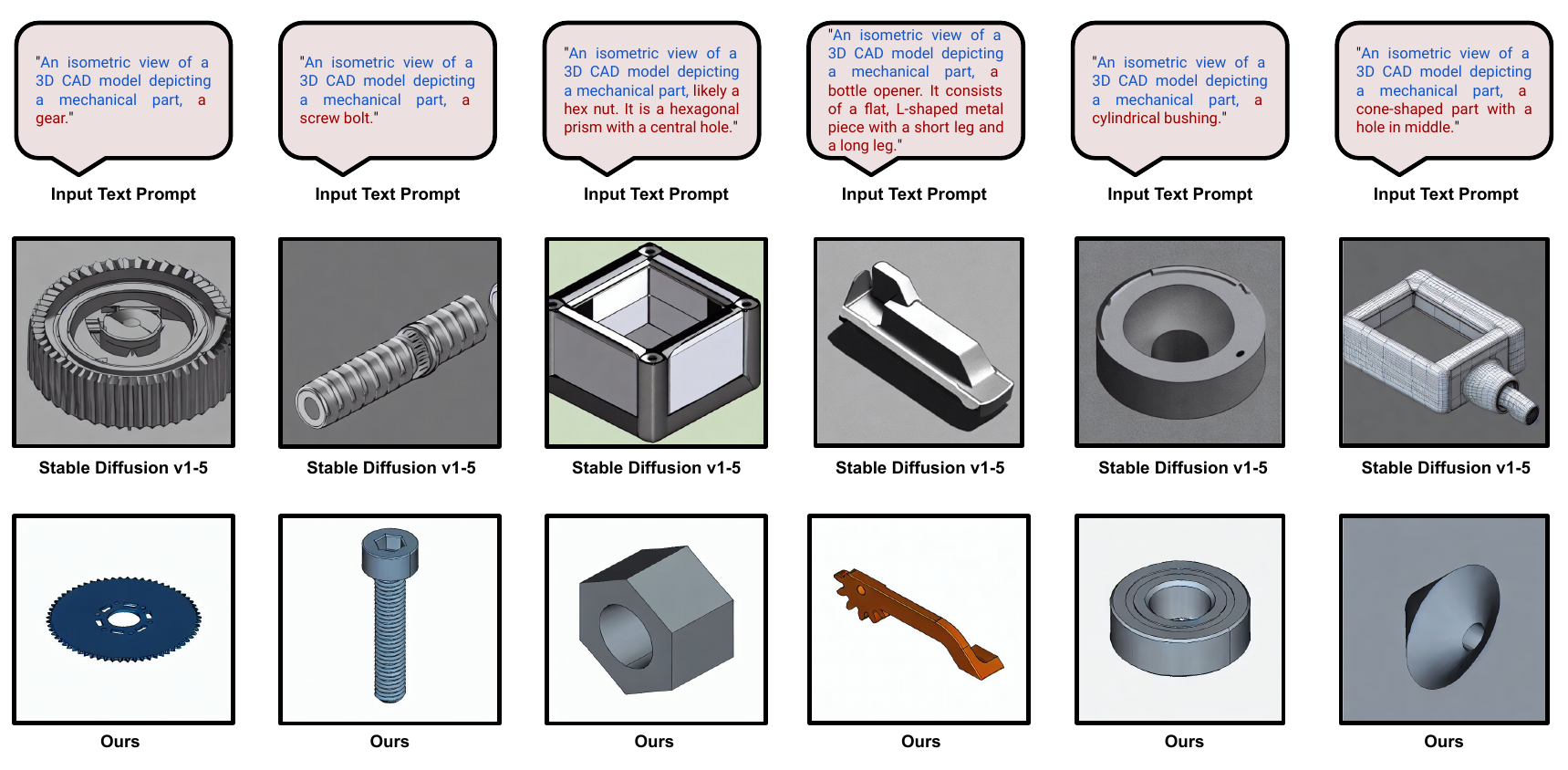}}
    \caption{\textbf{Generated isometric view image given a text prompt.} The images are generated by the fine-tuned Stable Diffusion v1-5 (ours) and its original version.}
    \label{fig:Text2CAD_txt2iso}
\end{figure}
\begin{table*}[h]
\small
\centering
\begin{tabular}{cccccc}
\hline
\textbf{Part} & \textbf{bevel gear} & \textbf{cylindrical flange} & \textbf{bracket} & \textbf{cylindrical pin} & \textbf{mounting plate} \\ 
\hline
\textbf{Feature 1} & hexagonal bore & multiple holes & T-shaped & flanged head & rounded edges\\
\hline
\textbf{\xmark} & {9\%} & {0\%} & {18\%} & {0\%} & {0\%} \\
\textbf{\cmark} & \textbf{94\%} & \textbf{100\%} & \textbf{68\%} & \textbf{90\%} & \textbf{100\%} \\
\hline
\textbf{Feature 2} & {angled teeth on a conical surface} & {a central hole} & {triangular} & {concentric through hole} & {central aperture}\\
\hline
\textbf{\xmark} & {8\%} & {5\%} & {0\%} & {0\%} & {0\%} \\
\textbf{\cmark} & \textbf{100\%} & \textbf{100\%} & \textbf{100\%} & \textbf{100\%} & \textbf{50\%} \\
\hline
\end{tabular}
    \caption{
        \textbf{Quantitative effects of including features.} The values are obtained by counting the part samples having the specific feature among 100 generated samples of the corresponding part.}
    \label{tab:Text2CAD_features}
\end{table*}

We fine-tune the Stable Diffusion v1-5 on the labeled isometric images generated by GPT-4 for $50,000$ training iterations. 
We use a batch size of $10$ and a resolution of $512\times 512$, with a fixed learning rate of $1\times10^{-5}$. 
Additionally, we fine-tune the Zero-1-to-3~\cite{Zero-1-to-3} model on the generated pairs of isometric images and their corresponding orthographic technical drawings, along with their relative camera poses. 
At each iteration, one of the top, front, or side technical drawings is randomly selected as the target image. 
This fine-tuning process is carried out for $10,000$ training iterations, using a batch size of $32$ and a resolution of $256\times 256$. 
We maintain a fixed learning rate of $1\times10^{-5}$.
All fine-tuning processes are conducted on 4 Quadro RTX 8000 GPUs to accelerate training and computation.
We will make the code and dataset publicly available upon the acceptance of the paper.

\subsection{Text to isometric image}
We qualitatively evaluate the fine-tuned Stable Diffusion model on our dataset compared to the pre-trained version using identical text prompts, as shown in Figure~\ref{fig:Text2CAD_txt2iso}. 
By visually inspecting the generated isometric images, we assess improvements in quality, detail, and fidelity to the prompts achieved through fine-tuning.

\begin{figure}[h]
\centering
\subfloat{
    \includegraphics[width=\linewidth]{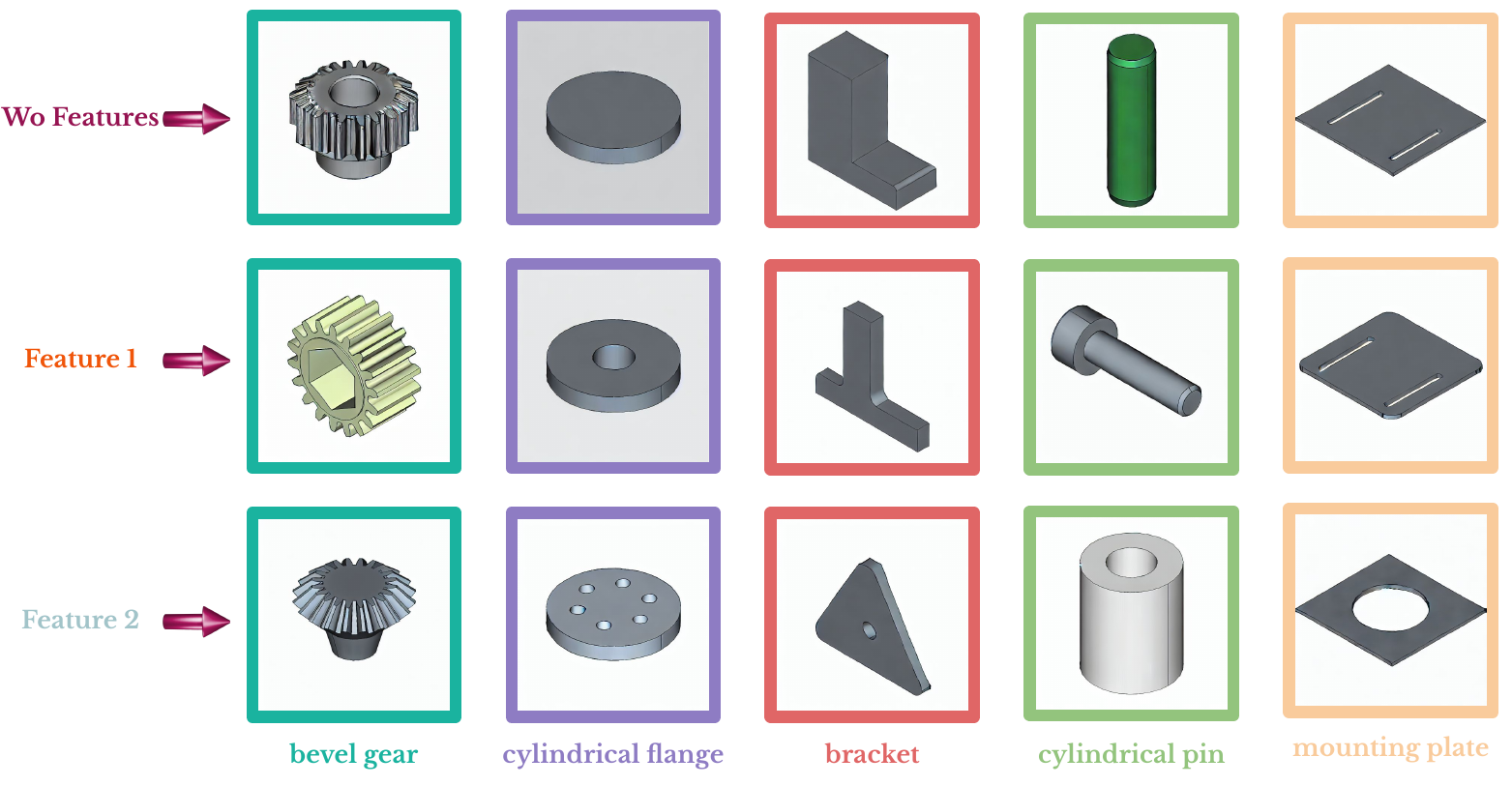}}
    \caption{
        \textbf{Qualitative effects of including features.}}
    \label{fig:Text2CAD_features}
\end{figure}

We also demonstrate how the fine-tuned model accurately generates images with specified features, as detailed in Table~\ref{tab:Text2CAD_features}. 
Generating 100 samples with and without the specified features, we found that the model significantly increases the presence of desired features when prompted. 
Figure~\ref{fig:Text2CAD_features} visually confirms this capability.

\begin{figure}[!h]
\centering
\subfloat[Evaluation by human]{
    \includegraphics[width=0.48\linewidth, page=1]{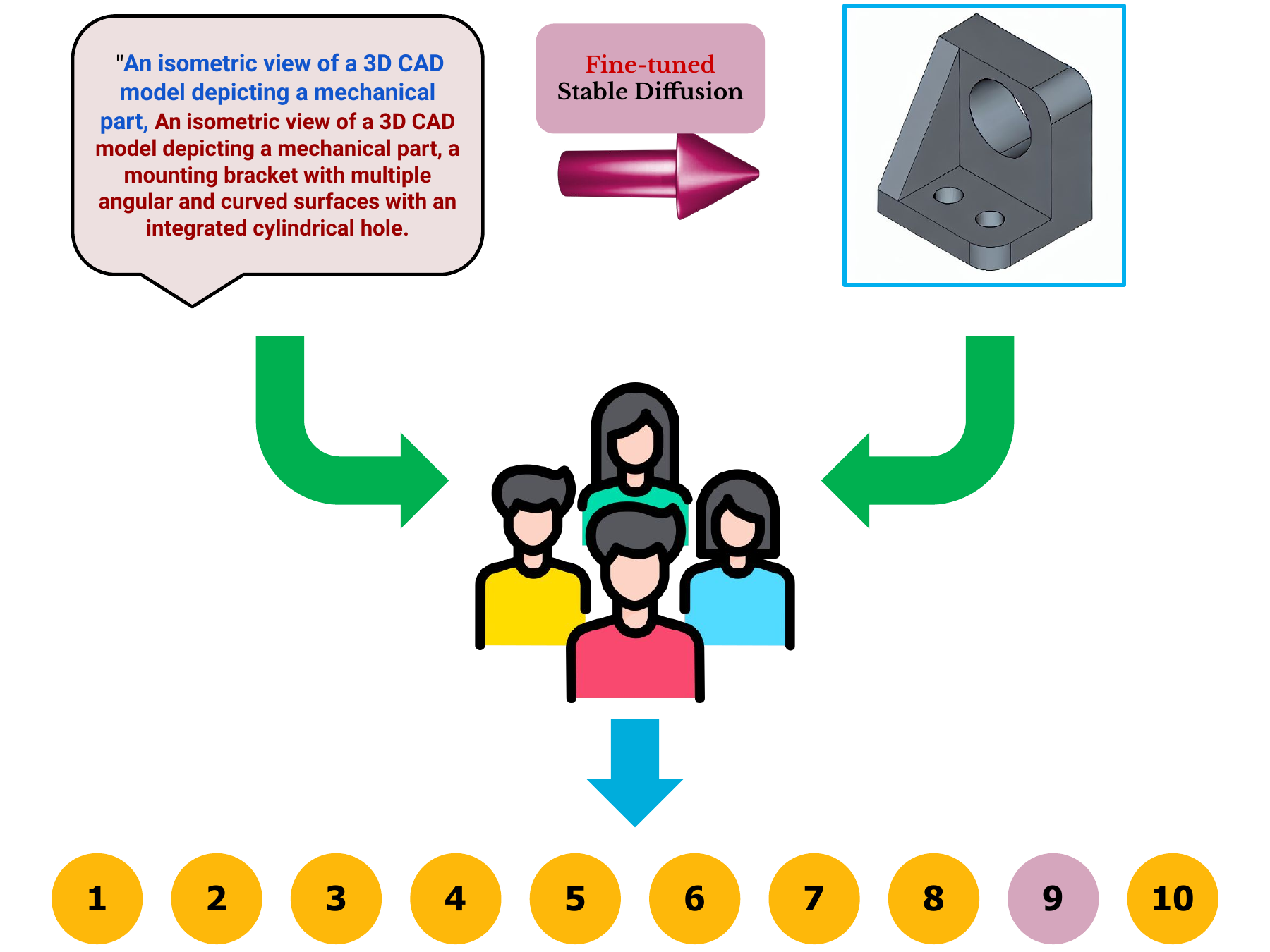}\label{fig:Text2CAD_mos_1}}
\hfill
\subfloat[Evaluation by GPT-4]{
\includegraphics[width=0.48\linewidth, page=2]{figures/Text2CAD_mos_.pdf}\label{fig:Text2CAD_mos_2}}
    \caption[Quantitative evaluation approaches]{\textbf{Quantitative evaluation approaches.}}
    \label{fig:Text2CAD_mos_a}
\end{figure}

\begin{figure}[!h]
\centering
\subfloat[Human vs. GPT-4]{
    \includegraphics[width=0.48\linewidth, page=3]{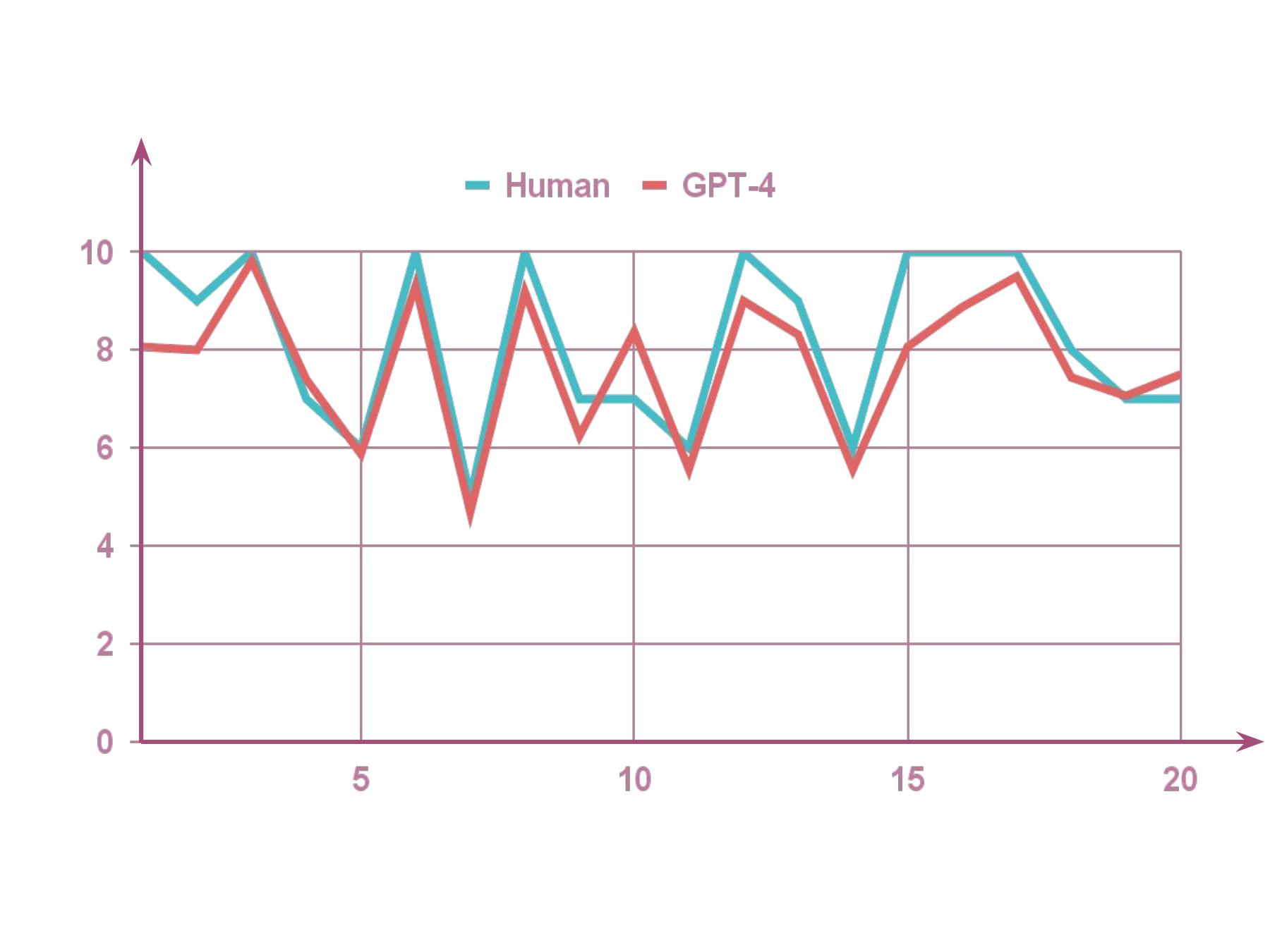}\label{fig:Text2CAD_mos_3}}
\hfill
\subfloat[Evaluation by GPT-4]{
\includegraphics[width=0.48\linewidth, page=4]{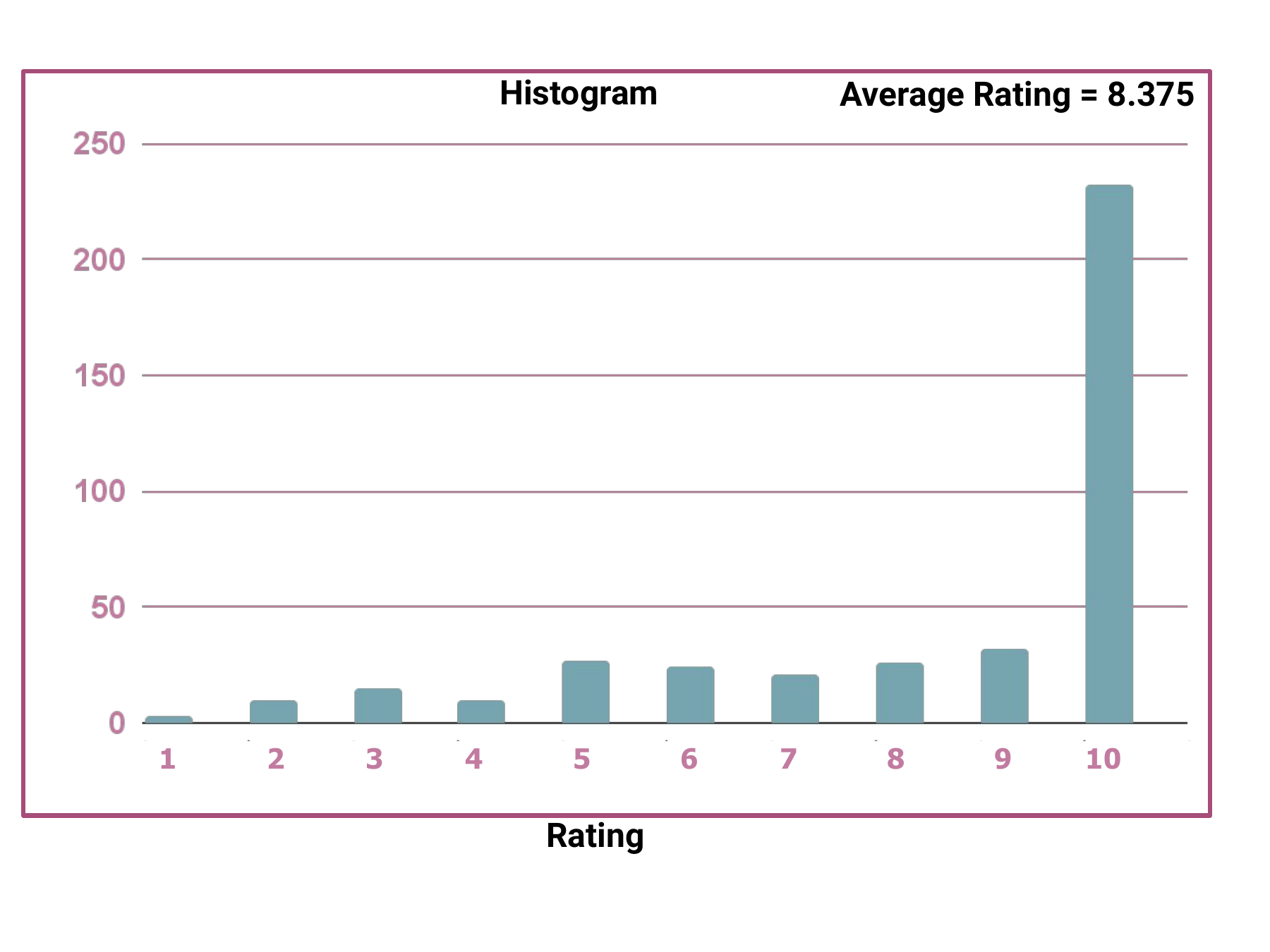}\label{fig:Text2CAD_mos_4}}
    \caption[Quantitative evaluations]{\textbf{Quantitative evaluations of isometric images.} The generated isometric images are evaluated by both human reviewers and GPT-4 for their accuracy in representing the given descriptions.}
    \label{fig:Text2CAD_mos_b}
\end{figure}

To quantitatively assess the fine-tuned model, we first have human evaluators rate the images based on their alignment with the descriptions, as shown in Figure~\ref{fig:Text2CAD_mos_1}. 
Given the difficulty of recruiting expert evaluators, we also use GPT-4 for comparison, as illustrated in Figure~\ref{fig:Text2CAD_mos_2}. 
Figure~\ref{fig:Text2CAD_mos_3} reveals a strong correlation between human and GPT-4 evaluations, validating GPT-4’s effectiveness in this role.

Furthermore, we task GPT-4 with generating textual descriptions for $100$ randomly selected objects following a predefined template. 
Subsequently, the fine-tuned Stable Diffusion model creates four images per prompt. 
Then, GPT-4 assesses the alignment between the generated images and their text descriptions.
As shown in Figure~\ref{fig:Text2CAD_mos_4}, most images received perfect ratings, with an overall average of $8.375$, demonstrating the robustness of our approach.

\subsection{Isometric to orthographic technical drawings}
\begin{figure}[!h]
\centering
\subfloat{
    \includegraphics[width=\linewidth]{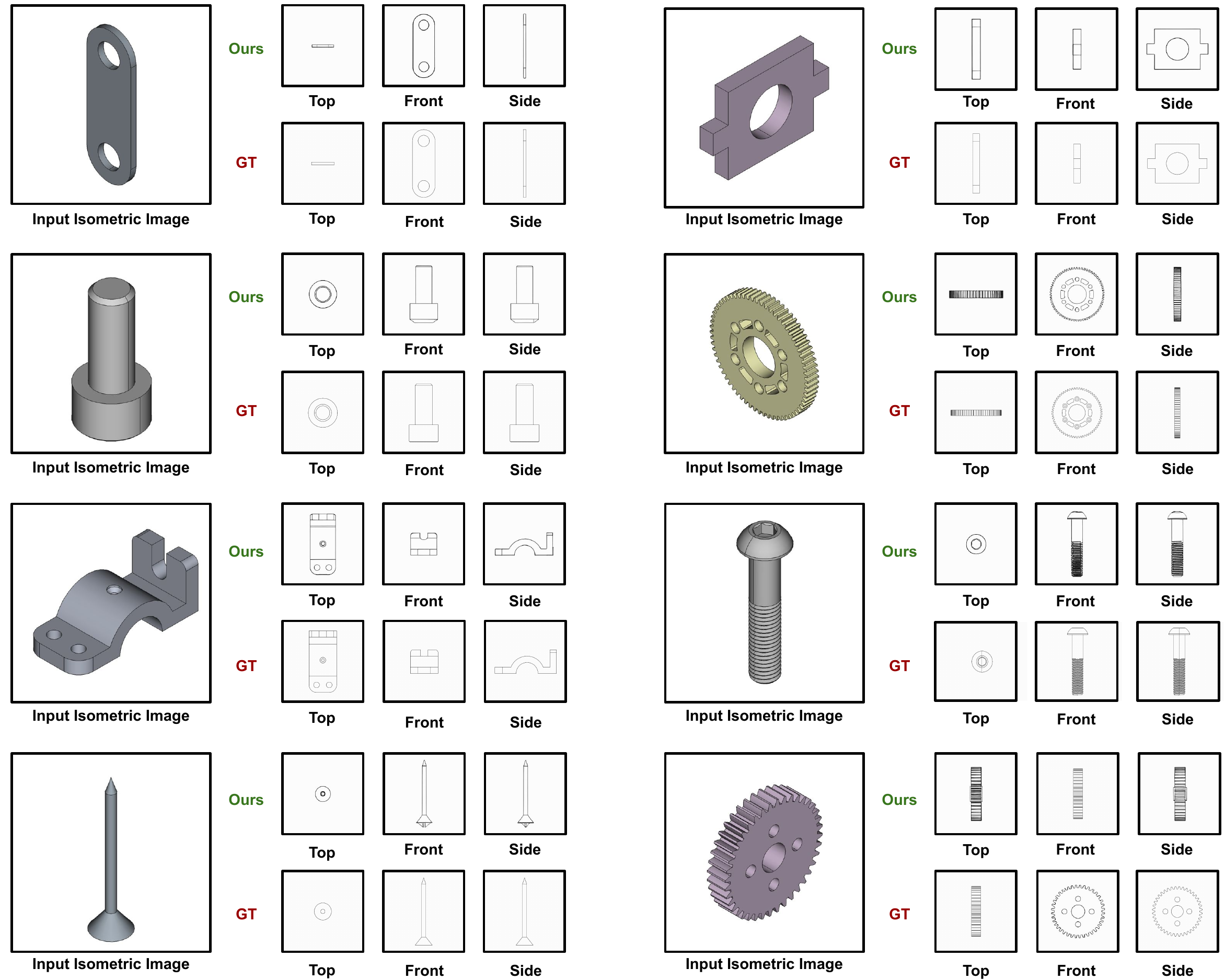}}
    \caption{\textbf{Orthographic technical drawings generation.}}
    \label{fig:Text2CAD_iso2ortho}
\end{figure}
We also visually evaluate the top, front, and side views of the technical drawings generated by the fine-tuned Zero-1-to-3~\cite{Zero-1-to-3} model against their ground truths, as shown in Figure~\ref{fig:Text2CAD_iso2ortho}. 
The results indicate that the model generates drawings with satisfactory detail.
\begin{table}[h]
\centering
\small
\begin{tabular}{lcccc}
        \toprule
        \textbf{View} &
        \textbf{Avg.$\downarrow$} & 
        \textbf{STD$\downarrow$} &        
        \textbf{Max$\downarrow$} & 
        \textbf{Min$\downarrow$} 
        \\
        \hline
        Top & {3.185} & {6.069} & {44.457} & {0.070}\\
        Front & {2.863} & {4.057} & {37.488} & {0.063}\\
        Side & {2.510} & {3.790} & {29.657} & {0.100}\\
        \hline
        Mean & {2.853} & {4.639} & {37.200} & {0.078}\\
        \hline
    \end{tabular}
    \caption{
        \textbf{Chamfer distance of technical drawings.}
    }
    \label{tab:Text2CAD_chamf}
\end{table}

To further assess accuracy, we measured the Chamfer distance~(CD) between the generated orthographic images and the ground truths. 
Table~\ref{tab:Text2CAD_chamf} shows the average, standard deviation, maximum, and minimum CD. 
Although direct comparisons with other methods aren't possible due to our approach's pioneering nature, the low average CD suggests high accuracy.

\subsection{Text to CAD}
\begin{figure}[!h]
\centering
\subfloat{
    \includegraphics[width=\linewidth]{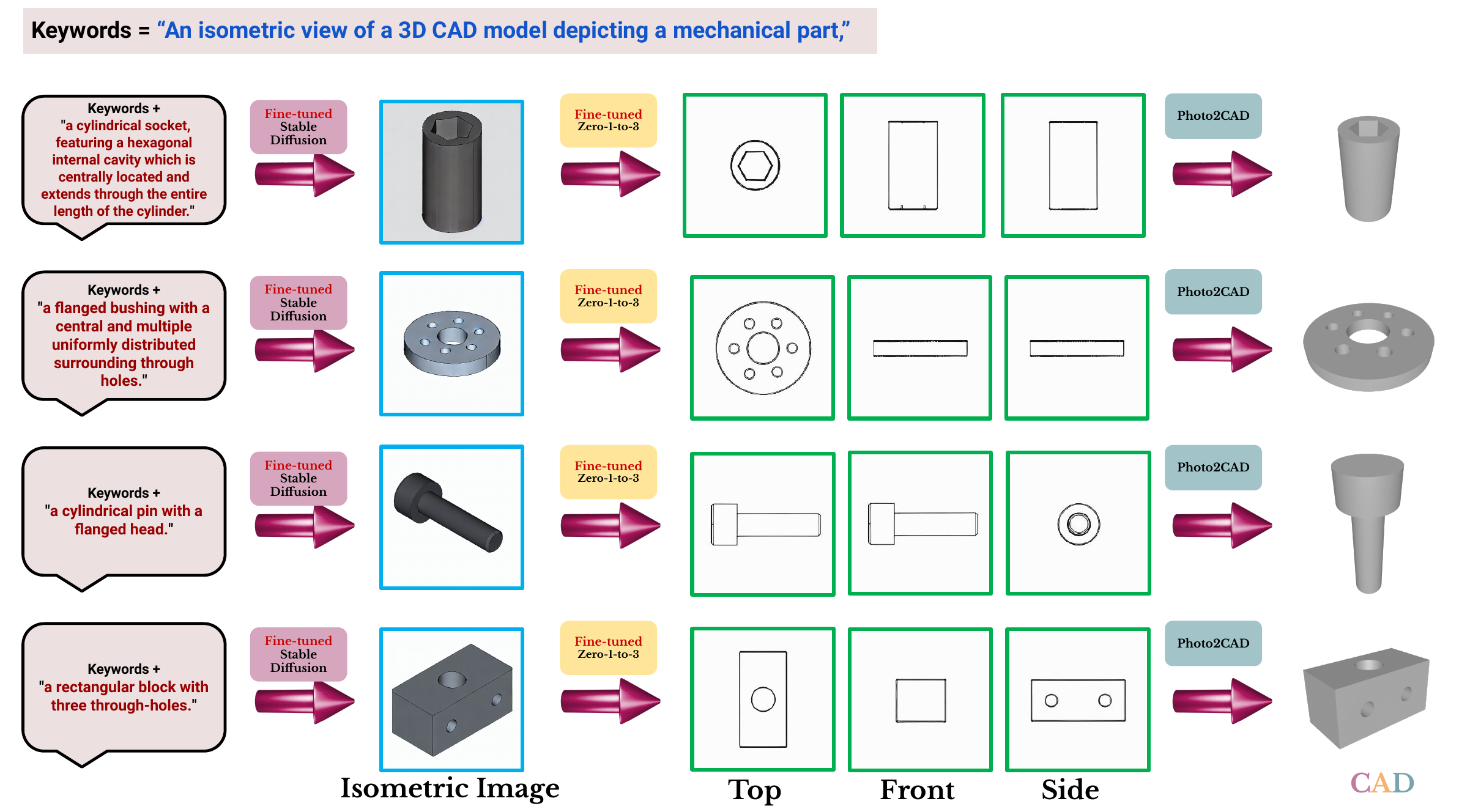}}
    \caption{\textbf{CAD generation from textual description.}}
    \label{fig:Text2CAD_txt2CAD}
\end{figure}

Figure~\ref{fig:Text2CAD_txt2CAD} illustrates our entire pipeline, including the final generated CAD model and intermediate steps like isometric images and orthographic technical drawings. 
This visualization highlights the effectiveness of our method in converting textual descriptions into detailed CAD models, demonstrating its potential to streamline the CAD design process in real-world applications.

\subsection{Ablation study}
In this section, we analyze our method through experiments and comparisons to assess the contributions of individual components to overall performance and effectiveness.

\paragraph{Diversity.}
\begin{figure}
\centering
\subfloat{
    \includegraphics[width=\linewidth]{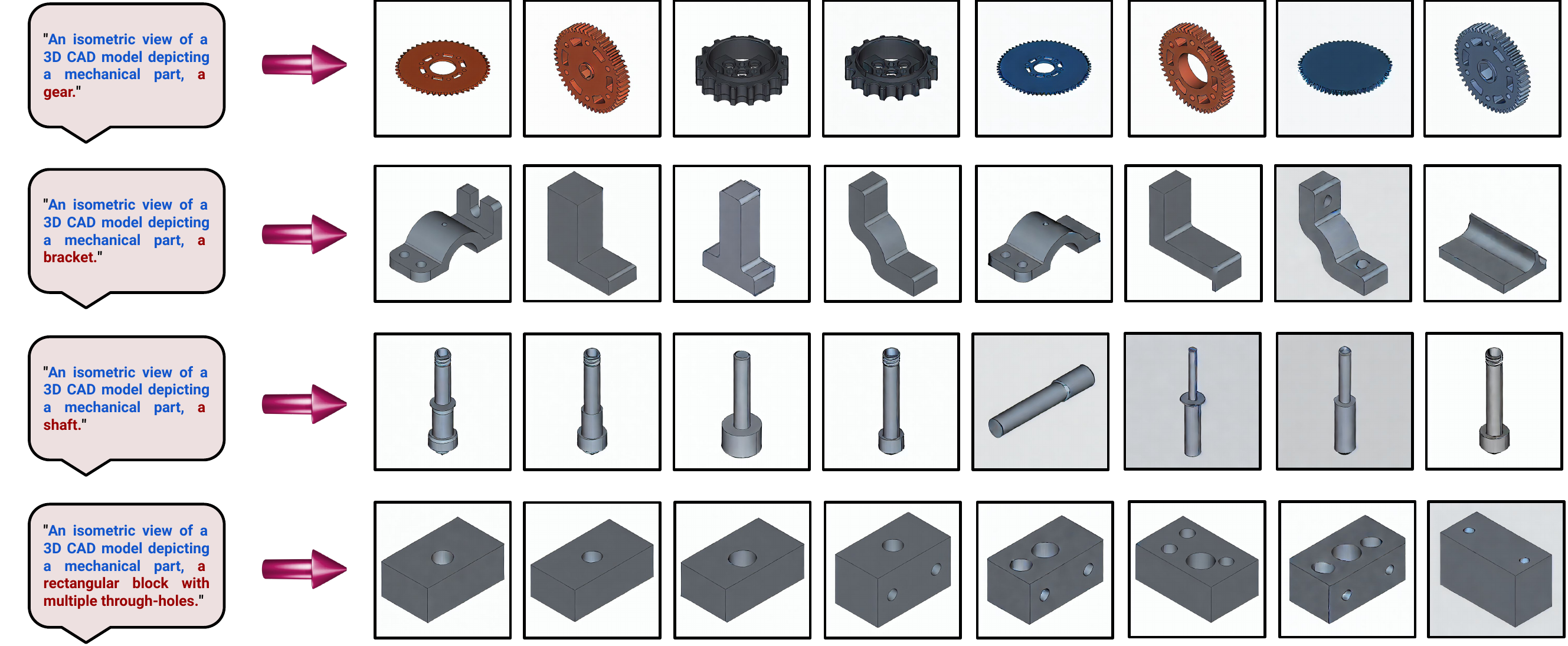}}
    \caption{\textbf{Diversity of generated shapes.}}
    \label{fig:Text2CAD_diversity}
\end{figure}
We evaluate the diversity of generated results for specific general prompts in Figure~\ref{fig:Text2CAD_diversity}. 
Our analysis reveals that despite fine-tuning the stable diffusion model on only a few samples from our dataset, it can generate a diverse set of images depicting various objects. 
This observation highlights the capability of our model to generalize to unseen data well and produce a wide range of outputs.

\paragraph{Orientation.}
\begin{figure}
\centering
\subfloat{
    \includegraphics[width=\linewidth]{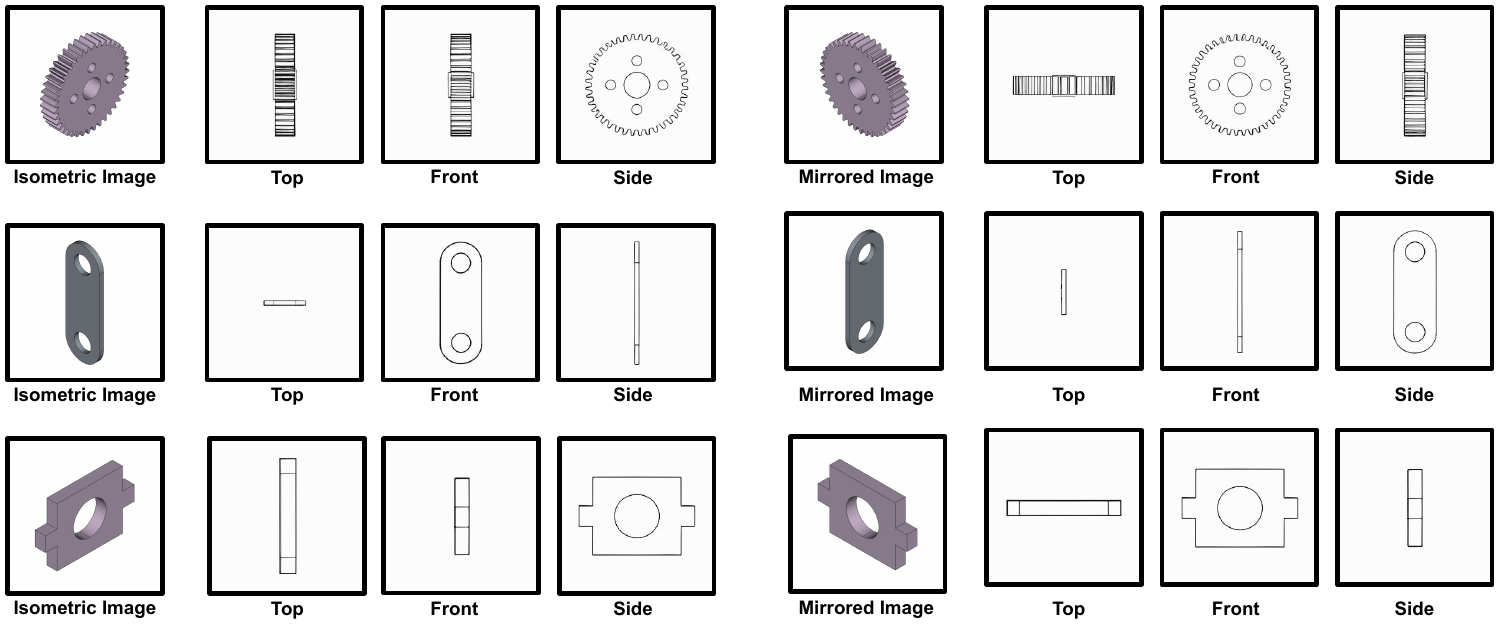}}
    \caption{\textbf{Effect of object orientation.}}
    \label{fig:Text2CAD_orientation}
\end{figure}
In this study, we assess how object orientation affects the generation of orthographic technical drawings from isometric images. 
Although the isometric viewpoint is fixed, varying the object's orientation impacts the resulting projections. 
We mirrored the isometric image to create different orientations and generated corresponding orthographic drawings for each. 
As shown in Figure~\ref{fig:Text2CAD_orientation}, our method consistently maps isometric images to the correct projections, regardless of orientation, with the side view of the original aligning with the front view of the mirrored images. 
This demonstrates the robustness of our approach.

\begin{figure}[!h]
\centering
\subfloat{
    \includegraphics[width=\linewidth]{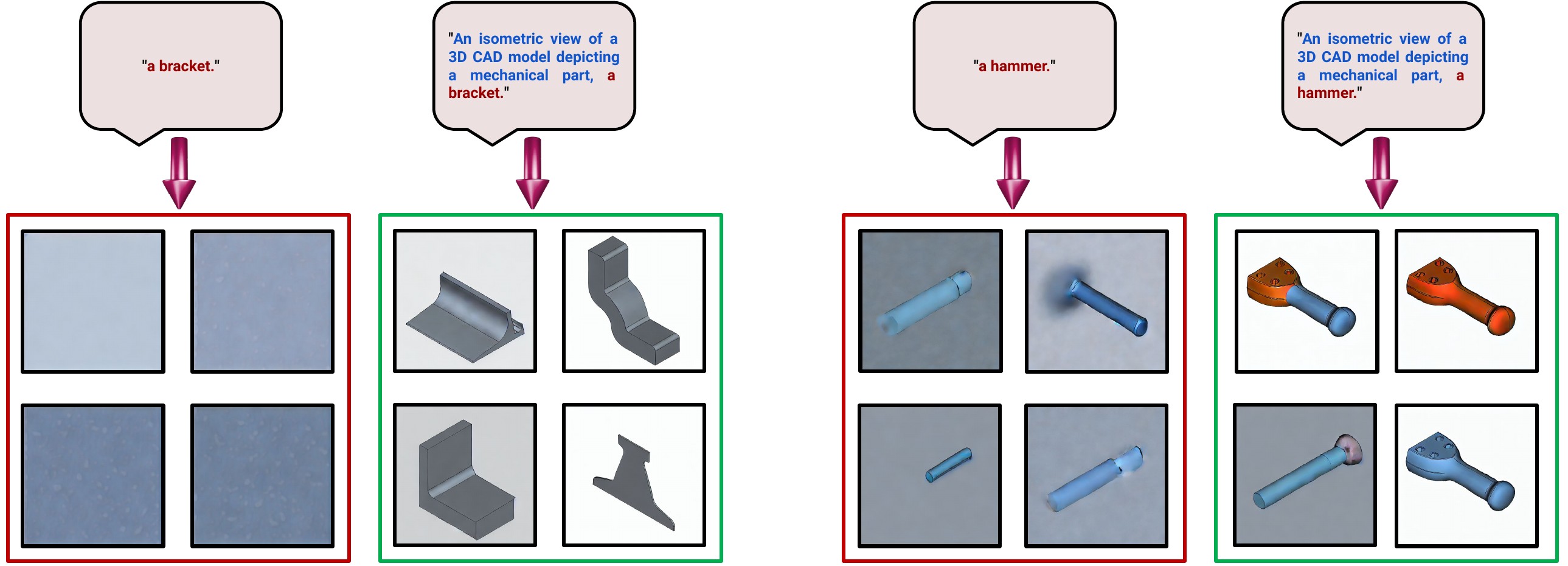}}
    \caption[Effect of keywords]{\textbf{Effect of keywords.}}
    \label{fig:Text2CAD_keywords}
\end{figure}

\paragraph{Keywords.}
As discussed in the dataset section, we consistently initiate all text prompts with the phrase {\it "An isometric view of a 3D CAD model depicting a mechanical part"} during the fine-tuning process. 
To evaluate the impact of these keywords on the generated results, we provide text prompts with and without these keywords and display the outcomes in Figure~\ref{fig:Text2CAD_keywords}. 
The findings underscore the significance of including these keywords to accurately generate isometric images of CAD objects, which are crucial for the subsequent isometric to orthographic transformation phase.

\section{Conclusion}
In conclusion, our paper demonstrates an effective method for generating CAD models from text prompts. We first use a fine-tuned stable diffusion model to create detailed images from isometric views. Then, we fine-tune a novel view generation model to produce orthographic technical drawings, which are used to create the CAD models.
Our experiments highlight the success of this approach in generating detailed CAD models from text. This advancement promises to accelerate design workflows, boost productivity, and enhance innovation in CAD modeling, making the design process more accessible and precise.

\bigskip

\bibliography{bib}

\end{document}